\documentclass{article}

\usepackage{microtype}
\usepackage{graphicx}
\usepackage{subfigure}
\usepackage{booktabs} % for professional tables
\usepackage{url}
\usepackage{amsfonts}
\usepackage{enumitem}

\usepackage{hyperref}

\usepackage[accepted]{icml2021}

\usepackage{amsmath}

\usepackage[suppress]{color-edits}
\addauthor{vs}{red}
\definecolor{burgundy}{rgb}{0.5, 0.0, 0.13}
\addauthor{emk}{burgundy}

\newcommand{\DoWhy}{{DoWhy} }

\newcommand{\dowhy}{dowhy}

\newcommand{\DoWhyNS}{DoWhy}
\icmltitlerunning{\DoWhyNS: Addressing Challenges in Expressing and Validating Causal Assumptions}

\begin{document}

\twocolumn[
\icmltitle{\DoWhyNS: Addressing Challenges in\\ Expressing and Validating Causal Assumptions}



\icmlsetsymbol{equal}{*}

\begin{icmlauthorlist}
\icmlauthor{Amit Sharma}{to}
\icmlauthor{Vasilis Syrgkanis}{to}
\icmlauthor{Cheng Zhang}{to}
\icmlauthor{Emre K\i c\i man}{to}
\end{icmlauthorlist}

\icmlaffiliation{to}{Microsoft Research}

\icmlcorrespondingauthor{Amit Sharma}{amshar@microsoft.com}

\icmlkeywords{\dowhy, sensitivity analysis, causal graph}

\vskip 0.3in
]



\printAffiliationsAndNotice{}%\icmlEqualContribution} % otherwise use the standard text.

\begin{abstract}
Estimation of causal effects involves crucial assumptions about the data-generating process, such as directionality of effect, presence of instrumental variables or mediators, and whether all relevant confounders are observed. Violation of any of these assumptions leads to significant error in the effect estimate. However, unlike cross-validation for predictive models, there is no global validator method for a causal estimate. As a result, expressing different causal assumptions formally and validating them (to the extent possible) becomes critical for any analysis. We present \DoWhyNS, a framework that allows explicit declaration of assumptions through a causal graph and provides multiple validation tests to check a subset of these assumptions. Our experience with \DoWhy  highlights a number of open questions for future research: developing new ways beyond causal graphs to express assumptions, the role of causal discovery in learning relevant parts of the graph, and developing validation tests that can better detect errors, both for average and conditional treatment effects. DoWhy is available at \url{https://github.com/microsoft/dowhy}.
%open questions
\end{abstract}

\section{Introduction}
Given observed data, causal inference is the process of using domain assumptions to estimate the effect of a desired action. Formally, a causal effect is always defined w.r.t. to a causal model (e.g., a structural causal model~\cite{pearl2009causality}) that encodes assumptions that typically cannot be learned from observed data. Therefore, in practice, causal inference involves a sequence of steps that start with specifying the necessary assumptions:  asserting domain assumptions, constructing a causal model or graph for the data-generating process based on the assumptions, identifying the desired effect based on the causal model, and estimating the effect.

In recent years, there has been substantial progress in better estimation methods for causal inference, in part by fusing machine learning algorithms with the traditional effect estimation methods~\cite{wager2018forest,Athey2019,pmlr-v70-hartford17a,chernozhukov2017double,shalit2017estimating,NIPS2017_6a508a60,Powers2018,lewis2018adversarial,syrgkanis2019machine,pmlr-v99-foster19c,NEURIPS2019_15d185ea,jacob2021cate,kennedy2020optimal,nie2021quasi,NEURIPS2020_8fcd9e54}. Specifically, in addition to estimating the average treatment effect (ATE), recent methods allow estimation of fine-grained conditional effects on based on given covariates, the conditional average causal effect (CATE). Given an observed dataset, many of these methods claim to estimate the causal effect of an intervention, even under heterogeneity of the effect across units, high-dimensional data, and non-linear effects.

The challenge, however, is that the prerequisites for the estimation task---building a causal model or graph for the problem---have not kept pace with the advances in estimation. As a result, advances in estimation do not immediately transfer to the practical process of causal inference because estimation methods assume that the causal graph has already been built and available. In practice, obtaining plausible assumptions to develop the causal graph either by human experts or by causal discovery algorithms is \vsedit{potentially} the biggest challenge of doing causal analysis and that is often overlooked in the recent causal machine learning literature. \vsedit{Moreover, educating users about causal machine learning algorithms (e.g., what are the causal graph assumptions that are implicitly being made by these techniques), so that they can construct the appropriate datasets where these assumptions are satisfied is the biggest challenge in adoption and correct use of these statistically powerful techniques.} In this paper, we highlight the fundamental importance of these assumptions, describe a framework for causal inference that helps analysts to express and test those assumptions, and propose open questions for future research.

\textbf{Assumptions are fundamental to causal inference.} Every causal estimate depends on assumptions that cannot be fully tested from observed data. Importantly, the bias due to these assumptions does not wash out even with infinite data.  Take for example, the common assumption for almost all CATE estimators,  that there is a set of \textit{known} confounders (common causes) of the treatment and outcome. What happens if one of the confounders is actually an instrumental variable or a mediator? If it is an instrumental variable, then conditioning on the instrumental variable will lead to a high variance estimate. In this case, the error is in the modelling  (or \textit{identification}) phase whereas it may appear as a problem with the  estimator's variance properties. If a mediator is incorrectly assumed as a confounder, the estimate will be biased but the error may not even be detectable. Finally, there  can always be unknown and unobserved confounders that can lead to errors.  
Sec. 2 shows simple examples on the difficulty of catching modelling versus estimation errors.

The end result is that we are left with advanced estimators with confidence intervals that assume that the modelling step is already correct, and thereby overestimate the certainty in their estimates. Importantly, there is no clear way of incorporating the uncertainty or bias due to errors in the modelling or identification phase. 
Moreover, there is no reliable method to validate an  estimate after the full causal analysis, unlike in supervised machine learning where cross-validation can determine the quality of any predictor.
In general, therefore, it is difficult to judge the quality of any obtained causal estimate from observed data, since the ``correct'' causal effect depends on the modelling assumptions.

\textbf{\DoWhyNS: Expressing and validating assumptions.} \DoWhy is a popular open-source python library for causal inference, having more than 300K downloads and used across many scenarios and fields.  Sec. 3 discusses how \DoWhy is designed to make assumptions  ``first-class'' citizens of a causal analysis. Its API implements causal inference in four steps: \textit{model}, \textit{identify}, \textit{estimate}, and \textit{validate}. The first two steps focus on expressing causal assumptions transparently using a causal graph while the last step provides multiple methods to  validate the resultant effect estimator based on the assumptions. The API structure is designed to help understand the interplay of the different steps in analysis. For example, to evaluate robustness of the estimate, one may try the same estimator with different causal models or identification strategies; or the same model with different estimators and validation tests to find the best estimator for a dataset. 

\textbf{Open questions.}
In Sec. 4, we describe the practical challenges in applying causal inference through our experience in working with users of \DoWhyNS. One of the biggest challenges is that a causal graph is often difficult to obtain  and there is a need for developing alternative abstractions for inputting causal assumptions. Moreover, there are fundamental challenges to building robust validation tests using observed data and  interpreting their results. 

\begin{figure}[t]
    \centering
    \subfigure[DGP: $w\rightarrow \{t, y\}; z \rightarrow t; t \rightarrow y$. $z$ is an instrument.]{
    \includegraphics[scale=0.3]{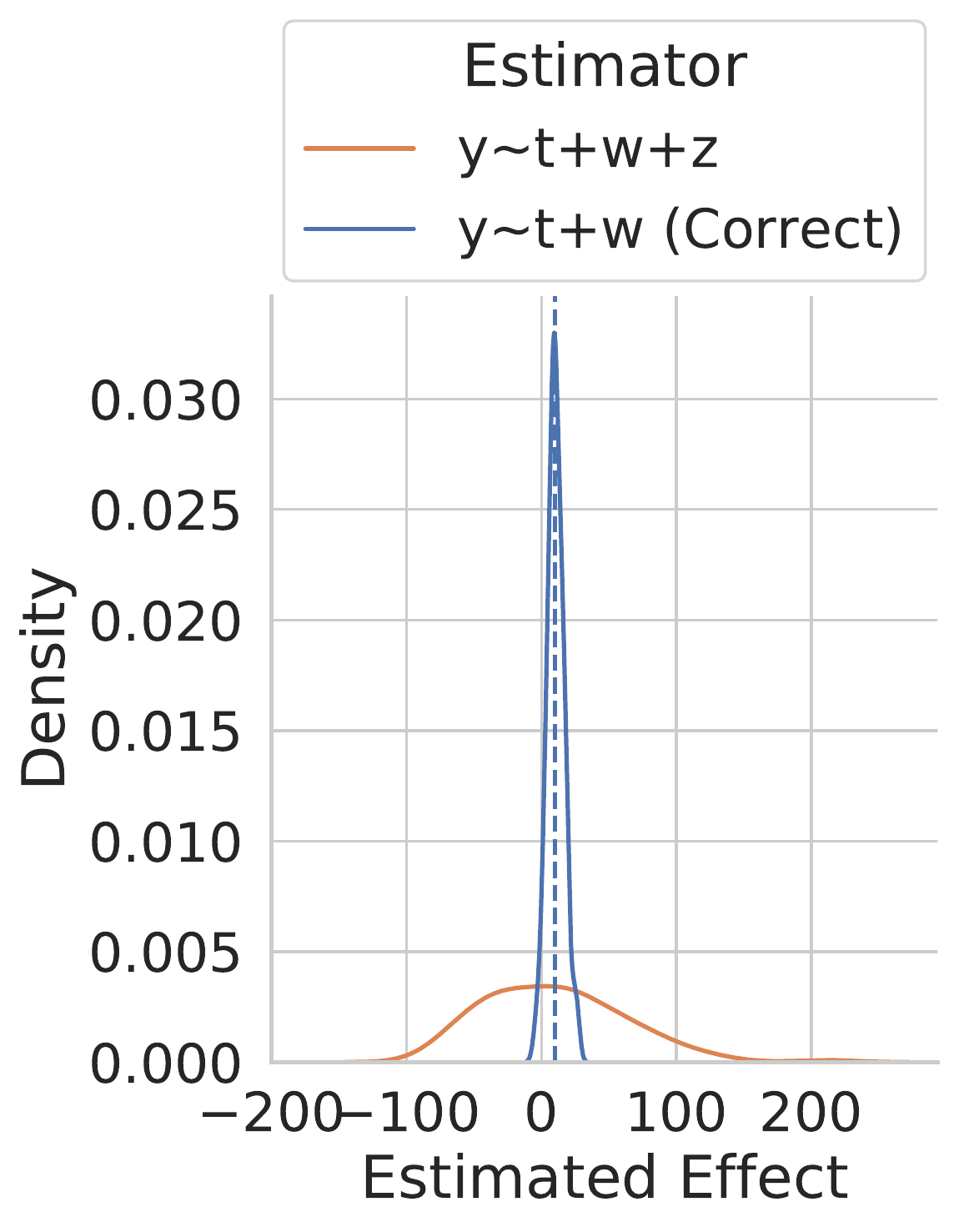}
    \label{fig:subfig1}
}~
   \subfigure[DGP: $ t \rightarrow m \rightarrow y$. $m$ is a mediator from $t$ to $y$.]{
    \includegraphics[scale=0.3]{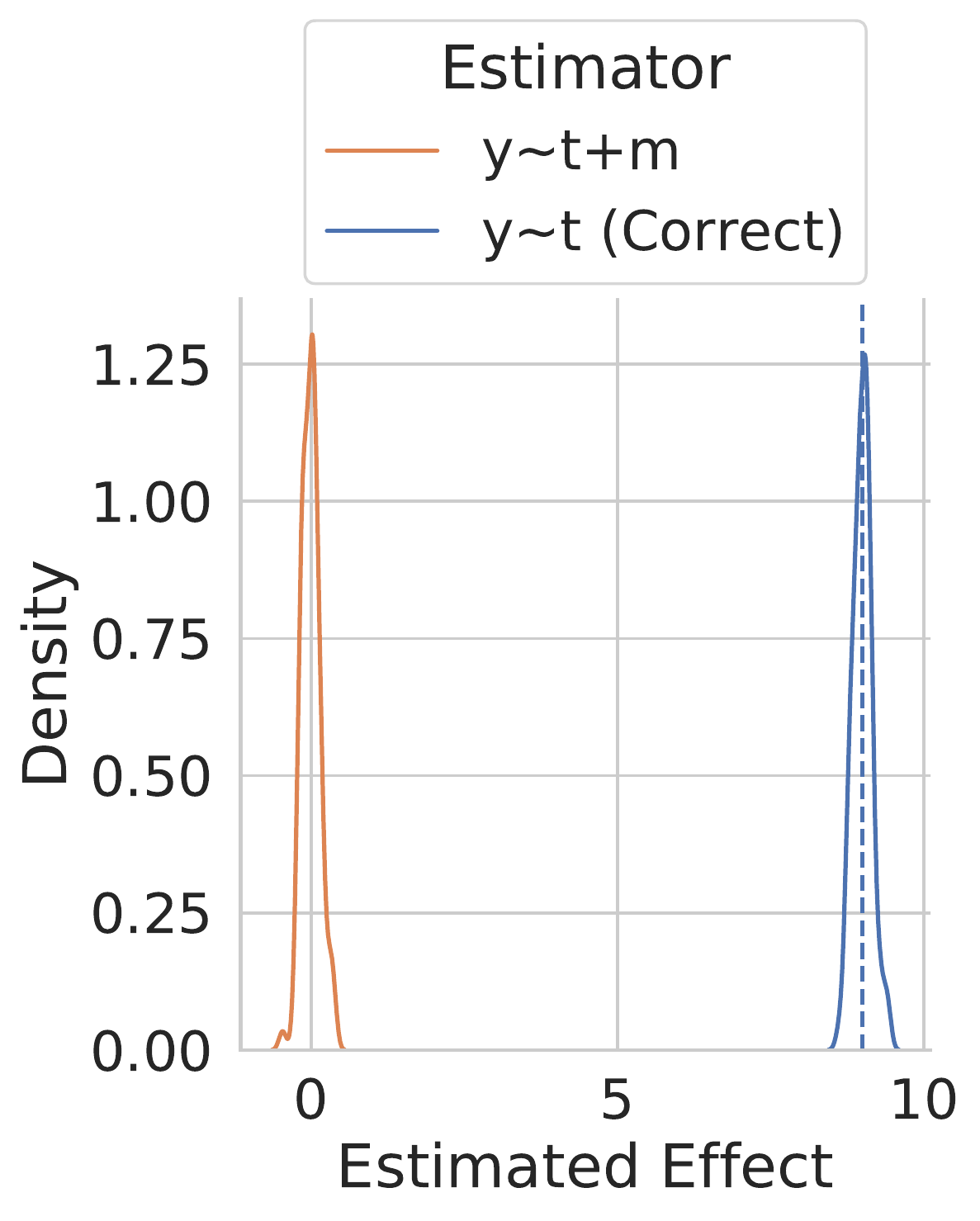}
    \label{fig:subfig2}
}
    \caption{Distribution of effect estimates under correct and incorrect identification. A data scientist works only with a  single dataset; however, to illustrate the distribution of the estimator, in each figure, we sample 100 datasets from the same data-generating process: blue lines show the correctly identified estimator, orange lines show the faulty estimator, and dotted line shows the true causal effect. }
    \label{fig:my_label}
\end{figure}

\section{The importance of modelling assumptions}
Biases in estimation due to unobserved confounders are well-known~\cite{greenland2003quantifying,robins2000sensitivity} and can  even change the sign of an estimated effect. While unobserved variables are a significant challenge to any causal analysis, estimation can go wrong even when all variables are observed. To complicate matters, it is usually difficult to even detect the error. And when errors are detected, it is difficult to know \textit{why}: is it a modeling/identification error or an estimation error?

To illustrate the difficulty of detecting errors in causal estimation, we present two simple examples. 
The first example shows a faulty estimator that leads to high variance estimate. It is possible to detect such high variance, but hard to determine the reason for it. The second example shows an estimator with bias where it is not even possible to detect that the effect estimate is incorrect. 
In particular, these estimators cannot recover the average treatment effect (ATE) over all units, which is a simpler problem than estimating the heterogeneous  conditional average treatment effects (CATE) that most recent work aims to estimate.

Throughout this paper, the goal is to estimate causal effect of a treatment variable $t$ on the outcome $y$, $\mathbb{E}[y|do(t=1)] - \mathbb{E}[y|do(t=0)]$. As mentioned above, we consider the simplest case where all variables in the causal graph are observed and relationship of $y$ with other variables is linear. 

\subsection{High variance estimate due to instrument}
Consider a dataset with four variables: treatment $t$, outcome $y$ and two covariates ($z$ and $w$). Using the backdoor criterion, a data scientist conditions on $z$ and $w$ to estimate the effect of $t$ on $y$. In this case, they use linear regression since it is known that the functional form for $y$ is linear. However, they observe that the resultant estimate has a high variance, as depicted in Fig.~\ref{fig:subfig1}. High-variance can be detected, for example, by creating multiple bootstrapped datasets and repeating estimation on each of them. To decrease variance, data scientists can apply more advanced estimation methods or can collect more data.

The above solutions presume that the problem is due to the estimator, but it is equally possible that the problem is in the \textit{identification} of effect. What if one of the covariates is an instrumental variable instead of a common cause? If that's the case, then the correct remedy is to remove that variable from the list of confounders and proceed as usual with the backdoor estimation. As it turns out, $z$ is an instrumental variable in the dataset for Fig.~\ref{fig:subfig1} (see DGP in Appendix)   and therefore the correct estimator is to condition on only $w$, as shown by the blue line. 
It is not possible to distinguish from observed data whether the estimation or identification step step causes the high variance. In practice, we need domain knowledge or extra assumptions to rule out identification flaws before dealing with estimation issues. 

\subsection{High bias estimate due to mediator}
Consider the same data scientist having a different dataset, treatment $t$, outcome $y$, and a variable $m$. Suspecting $m$ to be a common cause confounder, they condition on $m$ and obtain an effect estimate of zero, as shown in Fig.~\ref{fig:subfig2}. The estimate is precise with small confidence intervals and one may be tempted to conclude that the treatment has no effect. 

But what if the variable is a mediator on the path from $t$ and $y$, and not a confounder? In that case, the estimate is biased (assuming that the goal is to estimate the total effect of treatment (which is often the case) and not just the direct effect). The true estimate is obtained without conditioning, as shown by blue line in Fig.~\ref{fig:subfig2} (effect estimate of 10). For an analyst relying on the estimate's precision on the observed data, there is no way to know that they have made an error in their analysis. Any remedy for such errors can only come from assumptions that the data scientist brings from external knowledge.  No amount of estimation tuning is going to fix this, and one will never know the error (unless there is a randomized experiment to confirm the effect). 
Finally, as mentioned above, these problems exacerbate in the presence of unobserved confounders.

\section{\DoWhy framework}
The above problems occur due to the lack of a reliable validation mechanism for causal estimates, unlike in supervised prediction tasks where one can use cross-validation to detect the quality of a predictive model and thereby ascertain whether the model's assumptions were plausible. In the absence of a global validator to choose among  causal estimators, there are two key challenges:
\begin{enumerate}[noitemsep,topsep=0pt,parsep=0pt,partopsep=0pt]
    \item \textit{What are the right causal assumptions? } The practice of modeling assumptions (i.e., translating domain knowledge into a causal graph) and the implications of these assumptions for identification and estimation.
    \item \textit{How to check those assumptions?} The practice of validating causal assumptions to the extent possible through tests and sensitivity analysis, since observational causal inference tasks has no ground-truth data available to evaluate on.
\end{enumerate} 

We believe that the right abstraction for causal analysis can help lay equal focus on these two questions, in addition to estimation. To this end,  we discuss \DoWhyNS, an end-to-end library for causal inference, that organizes causal analysis around the  four key steps:  Model, Identify, Estimate,  and Refute/Validate. 
 \textbf{Model} encodes prior knowledge as a formal causal graph, \textbf{identify} uses graph-based methods to identify the causal effect, \textbf{estimate} uses statistical methods for estimating the identified estimand, and finally \textbf{refute} tries to refute the obtained estimate by testing robustness to initial model's assumptions.

\begin{footnotesize}
\begin{verbatim}
model = CausalModel(data, graph, 
    treatment, outcome)
estimand = model.identify_effect()
estimate = model.estimate_effect(estimand,
 method_name="propensity_score_weighting")
refute = model.refute_estimate(estimand,
    estimate,
    method_name="placebo_treatment_refuter")
\end{verbatim} 
\end{footnotesize}

The focus on all the four steps, going from data to the final causal estimate (along with a measure of its robustness) allows a user to formally express and test causal assumptions. None of the existing libraries for causal inference support an easy way to conduct multiple refutations, even though sensitivity and robustness checks are an important part of causal analysis. 
Most libraries in Python and R focus only on one of the steps.  For example, pcalg~\cite{kalisch2012pcalg} and dagitty~\cite{textor2016dagitty} focus on modeling; causaleffect~\cite{tikka2017causaleffect} on identification; and EconML~\cite{econml} and  CausalML~\cite{chen2020causalml} on estimation. While the Ananke library~\cite{ananke} has support for modelling, identification and estimation, it supports a limited set of estimation methods. Instead, \DoWhy is designed as a general API framework that can work with externally available implementations of methods for each step. For example,  \DoWhy integrates seamlessly with EconML and CausalML packages and allows calling any of their estimation methods (e.g., \texttt{method\_name=`econml.dml.DML'} for the double ML estimator~\cite{chernozhukov2017double}). 

Below we discuss how \DoWhyNS's API addresses the two challenges described above: the expression of causal assumptions, and the validation of causal estimators.

\subsection{Expressing causal assumptions}
Based on the seminal work of Pearl~\cite{pearl2009causality} in expressing a causal model through graphs, causal graphs have emerged as a popular abstraction for representing domain assumptions. Besides their presentational simplicity, causal graphs are also backed by do-calculus~\cite{pearl2012calculus}, a set of rules  and associated algorithms for identifying the causal effect in any given causal graph.

Before starting any causal analysis, \DoWhy stipulates that the user provide a causal graph over the observed variables. Drawing the causal graph encourages a user to think about the relationships between variables and determine the type for different variables (e.g., are they a confounder, instrumental variable, or a mediator?). 
Based on the causal graph, \DoWhy uses graph-based criteria and do-calculus to find expressions that can identify the causal effect. That is, it automatically identifies the causal effect (including mediation effects) using different identification algorithms, including
Back-door,
 Front-door criterion~\cite{pearl2009causality},
Instrumental Variables,
and the ID algorithm~\cite{shpitser2008complete}. 

Given that identification is automatic given a causal graph, the focus is on obtaining the causal graph  and making sure that it expresses the desired causal assumptions. The benefit of expressing assumptions formally is that their effect on the downstream estimation tasks can be quantified easily. For example, a user may construct multiple versions of the graph based on their domain knowledge, and then rerun the estimation method for these graphs to observe how different causal assumptions affect the final estimate. In the example discussed in Sec. 2.2, a user may supply two causal graphs: one with $m$ as a confounder and one with $m$ as a mediator. Both these graphs can be used to identify the effect and construct an estimator, and the differences in the resultant estimates will help understand the significance of the different causal assumptions for the particular dataset.

\subsection{Validating causal assumptions and estimators}
While causal graphs provide a formal abstraction for expressing causal assumptions, no such well-accepted method exists for validating those assumptions. Most work on testing causal assumptions focuses on \textit{sensitivity analyses}~\cite{rosenbaum2014sensitivity,robins2000sensitivity,veitch2020sense} that describe how an obtained estimate changes if we change any of the identifying assumptions. An analyst can use the sensitivity of the estimate to ascertain its robustness, perhaps based on plausible values of the assumption violations. However,  a key limitation is the qualitative nature of such analyses that prevent development of a more general test. 

Recent work on developing validation tests for causal estimators can be divided into two types. Similar to the cross-validation loss metric, the first type of \vsedit{methods~\cite{alaa2019validating,schuler2018comparison,Powers2018,athey2019estimating,nie2021quasi,pmlr-v99-foster19c,dwivedi2020stable} use observed data} to create a metric that denotes the quality of an obtained causal estimate. \vsedit{However, in the absence of ground-truth data with the true causal effects, and in the absence of experimental data from a randomized-controlled trial,} these methods use secondary estimators to 
estimate auxiliary predictive models that are being used in the quality metric. The validity of these metrics as evaluators of the quality of the causal effect, depends on the quality of these auxiliary estimators. Even though many of these techniques argue that the estimation error of these auxiliary models has a second order impact on the validity of the metric, this is only the case if these auxiliary estimators are already quite good and importantly are non-parametrically consistent. Achieving these requirement can be sometimes hard, if not harder, than the task of fitting the final causal model.

The second type of methods~\cite{neal2020realcause} construct a new, simulated dataset where the full data-generating process (DGP) is known  and hence the ground-truth  causal effect is also known. Candidate estimators are evaluated on their error on the new dataset, hoping that the extent of errors on the new dataset will transfer to the original dataset too. The challenge here is to create a simulated dataset that is \textit{close} enough to the original dataset, but yet it is possible to ascertain relevant parts of its  DGP to know the ground-truth causal effect. Different simulated datasets may give a different ranking of candidate estimators. 

Acknowledging the limitations in developing a ground-truth auxiliary metric from the first type of methods and the strong assumptions therein, \DoWhy provides a suite of validation tests based on creating simulated datasets similar to the original one, where the ground-truth causal effect is already known. It also provides sensitivity analyses to help evaluate causal estimators in conjunction with domain knowledge. It is important to note here that causal assumptions cannot be fully verified. Rather, the intent is to validate some necessary conditions entailed from a given assumption, and thereby filter out the models that do not satisfy them.  
Below we outline the supported validation tests. 

\textbf{Replacing treatment or outcome.} These tests replace the treatment or outcome variable to create a new dataset where the causal effect is trivially known. These are the closest to opaque-box cross-validation for predictive models---they can detect errors but not reveal which of the assumptions is violated. These tests are global in the sense that they can detect error due to any part of the analysis: incorrect identification, estimator, or even implementation bugs. 
\begin{itemize}[noitemsep,topsep=0pt,parsep=0pt,partopsep=0pt]
\item \textbf{Placebo Treatment}: When we replace the true treatment variable with an independent random variable, the estimated causal effect should go to zero.
\item \textbf{Dummy Outcome:} When we replace the true outcome variable with an independent random variable, the estimated causal effect should go to zero.
\item \textbf{Simulated Outcome~\cite{neal2020realcause}}: When we replace the outcome with a simulated outcome based on a known data-generating process close to the given dataset, the estimated causal effect should match the effect parameter from the data-generating process.
\end{itemize}

\textbf{Adding ``unobserved'' confounders.} These methods add random confounders where it is known that the effect should not change, or introduce correlated confounders to study the sensitivity of an estimate to unobserved confounding.
\begin{itemize}[noitemsep,topsep=0pt,parsep=0pt,partopsep=0pt]
\item \textbf{Add Random Common Cause:}  Adding a synthetic independent random variable as a common cause should not affect the estimated causal effect.
\item \textbf{Add Unobserved Common Causes: }  The effect estimate should not be too sensitive to additions of a common cause (confounder) to the dataset that is correlated with the treatment and the outcome?
\end{itemize}

\textbf{Creating subsets of the dataset.} These methods use subsets or bootstrap samples to check variance of the estimator.
\begin{itemize}[noitemsep,topsep=0pt,parsep=0pt,partopsep=0pt]
\item \textbf{Data Subsets Validation:} When we replace the given dataset with a randomly selected subset, the estimated effect should not change significantly.
\item \textbf{Bootstrap Validation}: When we replace the dataset with bootstrapped samples from the same dataset, the estimated effect should not change significantly.
\end{itemize}

Many of the above methods aim to refute the full causal analysis, including modeling, identification and estimation (as in \textit{Placebo Treatment} or \textit{Dummy Outcome}) whereas others refute a specific step (e.g., \textit{Data Subsets} and \textit{Bootstrap} only test the estimation step). Borrowing terminology from the software testing literature, the former can be called  ``integration'' tests whereas the latter can be called ``unit'' tests.

\section{Open research questions}

Through our work and the many users of \DoWhyNS, we have improved our understanding of the challenges people face while applying causal methods.  Here we describe three of the most critical fundamental research challenges.

\subsection{Eliciting assumptions from experts}

\vsedit{Causal methods require causal domain knowledge, expressed in the form of a Directed Acyclic Graph (DAG) or a Structural Causal Model (SCM), to guide the causal reasoning task.  Despite the seeming user-friendliness of DAGs,} we find it is often difficult to elicit such knowledge from domain experts. 
Often, it is difficult for domain experts to appropriately circumscribe the causal factors relevant to their study, separating exogenous from endogenous factors, and capturing key relationships at an appropriate abstraction.  Sometimes, domain experts do not understand core causal concepts, confusing correlational with causal relationships.
Perhaps most challenging is when even domain experts do not have a sufficient understanding of the given system.  
\vsedit{Especially when a large set of variables is included in the dataset, it is a daunting and cumbersome task to define a full causal model, capturing the relationship among a potentially high-dimensional set of variables.} 
There are several promising possibilities for addressing these issues, by allowing human input in a different format than graphs. 

\textbf{Machine teaching.}
    {\em Machine teaching methods} developed to elicit domain knowledge for conventional machine learning models  may be adaptable to elicit causal domain knowledge~\cite{simard2017machine}.  Such a human-in-the-loop approach may allow experts to refine their assumptions through iterative probing.  For example, a system might guide experts by presenting them with what-if scenarios, data-driven feedback, or other mechanisms to solicit confounders, mediators, and other common causal structures.

\textbf{Expressing assumptions as constraints.}
Often people may not be able to express causal relationships as edges between variables, but provide constraints that the causal graph should satisfy. The constraint may be  pair-wise independence constraints,  no direct-effect constraint, monotonicity constraint (Effect of $A$ on $B$ is positive), and so on. These constraints do not necessarily lead to a unique graph and thus require modified identification and estimation methods.
%to deal with them. %partially specified graphs. 

\textbf{High-level causal graphs.}
For some tasks, we may be able to use minimal causal graph representations rather than a full description of a system. In domains like images or text, it is unlikely that a causal graph over the raw inputs would be meaningful.  
For example, in the computer vision task of object recognition, the causal relationships may be at level of high-level features object shape, color, and texture, rather than relationships expressible at the granularity of individual pixels. For prediction tasks, high-level causal graphs have been proposed that help enforce \vsedit{certain invariances~\cite{arjovsky2019invariant}.} It will be useful to extend such high-level graphs to effect inference tasks. 

\textbf{Sufficient partial graphs.} Moreover, in many application settings, there is a target outcome variable of interest and a known treatment variable and typically only a subset of the graph needs to be defined to devise an identification strategy (i.e., specifying confounders, instruments, mediators). Eliciting a sufficient partial causal graph will likely require both interaction and iteration with a domain expert, and new strategies for testing the validity threat posed by unidentified relationships in the context of a particular data set.

\subsection{Causal discovery methods}
{\em Causal discovery methods}~\cite{spirtes2000causation,peters2017elements, glymour2019review} may assist domain experts in identifying key structures, and for validating a given DAG's plausability in the context of specific data set.
In addition to domain knowledge, one can consider using causal discovery methods to determine the causal graph. 
However, the fields of causal discovery and inference have evolved separately and many challenges remain for causal discovery to be useful in the downstream effect inference task.

\textbf{Incorporating domain knowledge.}
In many applications, we may have partial domain knowledge. To obtain the most accurate causal graph, utilizing both domain knowledge and causal discovery methods would be critical. However, most causal discovery algorithms are designed without a trivial way to incorporate the domain knowledge. Thus, it is important to design an interactive system that allows users to express their partial graph, obtain the discovered graph, and then update it based on domain knowledge. We also need to build discovery algorithms that can work iteratively with partial domain knowledge. 

\textbf{Graph representations.}
Without  experimental data or additional assumptions, most discovery algorithms can only identify the causal graph up to Markov equivalent classes and the output is represented as, for example, complete partial DAG (CPDAG), acyclic directed mixed graphs (ADMG) and partial ancestral graphs (PAG) \cite{spirtes2000causation, peters2017elements,bhattacharya2021differentiable}. However, causal inference typically requires one DAG and needs to be extended to handle these formulations.

\textbf{Additional untestable assumptions.}
Use of causal discovery algorithms will include additional assumptions on the causal inference pipeline.  Markov condition and faithfulness assumptions are the most common ones. 
Functional causal models~\cite{shimizu2006linear,zhang2009identifiability,zhang2010distinguishing}, which can be used to discovery the unique DAG require further assumptions regarding  functional and noise form of the structured equations. It is unclear how these assumptions will interact with the assumptions of the downstream effect inference task, and how the errors may propagate. More work is needed on combining discovery and effect inference as a single task and determining the quality of estimators possible under the scenario, including concerns such as multiple testing due to different graph structures \vsedit{and invalidity of subsequent $p$-values from the estimation methods deployed on the discovered causal graph.}

\subsection{Validating causal assumptions}
Given the impossibility of a global validator metric for causal effect, there will always be multiple validation tests to choose from. How does one choose estimators when tests do not agree? Can we create better, more robust tests? % that can be accepted as standard?

\textbf{Interpreting violations of assumptions.}
We have found that understanding and interpreting the results of sensitivity analyses and refutation tests is not always straightforward.  For example, if a data scientists collects multiple observational data sets from a system, and a given refutation test fails on only one of them, should the data scientist trust results from any of the data sets?

A related question is to understand when can sensitivity analyses and validation tests help us choose between estimators. Are there specific dataset properties or DGP properties that make validation tests more conclusive?
Another direction is to design validation tests as debugging methods in the process of improving an estimator: a good estimator has the fewest failed validation tests. This has parallels in assessing robustness and out-of-distribution generalization of predictive models~\cite{checklists,dash2020counterfactual}, where perturbation and other stress tests have been proposed. 

\textbf{Building better validation tests.}
Finally, there is the fundamental question on the limits to validation testing from observed data. While there may be a natural limit on general validators, a promising direction is to develop specific validators based on known assumptions. For example, would validators be more powerful if we knew the direction of the confounding effect, monotonicity of relationships, or specific independence constraints (and that are known to be true)? \vsedit{Recent progress has been made in some of these directions \cite{Cinelli2020,jesson2021quantifying}, but there remains many opportunities for more domain-expert guided validation tests, to improve on their practical relevance.}

Given that randomized data is the gold-standard to evaluate a causal estimate, a second direction is to explore how much randomized data is needed. Can validation methods be used alongside randomized data, and reduce the amount of randomized data needed? \vsedit{Can we close the circle and automate the generation of candidate experiments that would validate an effect estimate, based on testing specific assumptions that a domain expert may flag?}\emkedit{~\cite{hyttinen2013experiment}}

\section{Conclusion}
We highlighted the importance of modeling assumptions in causal effect estimation, and how this provided the motivation for including of modeling and validation as first class stages in the \DoWhy API.  We see the improvement of modeling assumptions and their validation as a research area critical to broadening the reliable usage of causal methods in practice.  Based on our experiences, we highlight three areas of open research questions to improve the elicitation of assumptions from experts, the incorporation of causal discovery methods across the causal inference process, and opportunities to improve the validation of assumptions.

\bibliography{example_paper}
\bibliographystyle{icml2021}

\appendix
\section{DGP for the Motivating Examples}

\subsection{Example 1}
        %\begin{equation}
            \begin{align}
            y & \leftarrow 10 t + 10w + \epsilon \\
            t & \leftarrow (\operatorname{sigmoid}(2z-1+w)>=0.5, 1, 0)\\
            z & \sim Bernoulli(0.5) \\
            w & \sim Normal(0, 0.4) \\
            \epsilon & \sim Normal(0,100) 
            \end{align}

\subsection{Example 2}
    \begin{align}
    y & \leftarrow 10m + \epsilon \\
            t & \sim Bernoulli(0.5) \\
            m & \sim Bernoulli(0.95t + 0.05(1-t)) \\
            \epsilon & \sim Normal(0,1) 
    \end{align}
        
\end{document}